\documentclass[11pt]{article}

\usepackage{arxiv}

\usepackage{times}
\usepackage{latexsym}
\usepackage{natbib}

\usepackage[utf8]{inputenc}
\usepackage{url}
\usepackage[hidelinks]{hyperref} 

\usepackage{graphicx}
\graphicspath{{Figures}}
    \DeclareGraphicsExtensions{.png,.pdf}
\usepackage{subcaption}
\usepackage{multirow}
\usepackage{booktabs}

\usepackage{float}
\usepackage{longtable}

\usepackage{amsmath}

\title{\large Do LLMs differentiate linguistic traits internally?: A layer-wise analysis}
\author{Samhita Bollepally \\
  \texttt{samhita.bollepally@tamu.edu} \\
  \And
   Takashi Yamauchi \\
  \texttt{takashi-yamauchi@tamu.edu} \\
   \And
    Manju Muralee\\
    \texttt{manjumuralee@tamu.edu}\\
  \And
   	Motunrayo Adebimpe\\
   \texttt{maryadebimpe@tamu.edu}\\
}

\begin{document}
\maketitle
\begin{abstract}
Large language models often produce judgments that resemble human responses, but it remains unclear whether they internally capture linguistic nuances such as conventional versus idiomatic meaning. A layer-wise representational analysis was conducted on two equal-sized pre-trained models, Llama-3.2-1B and Gemma-3-1B-PT, to examine whether and where these models differentiate linguistic traits from input. The analysis combined layer-wise probing with geometric distance measures, using the top embedding dimensions ranked by principal component loadings for each layer. The results show that even with equal capacity, models organize linguistic information in fundamentally different ways, suggesting the need to analyze internal representational geometry, examining not just behavioral outputs but semantic space alignment when evaluating how closely human and model language interpretation correspond.
\end{abstract}

\section{Introduction}

Recent work on large language models (LLMs) increasingly investigates how linguistic and semantic information is distributed across layers, challenging the assumption that deeper or final layers necessarily provide the most informative representations for downstream tasks. Layer-wise analysis enables tracking of how category-relevant structure emerges, stabilizes, or dissipates across layers. While early accounts proposed monotonic improvements in representational quality under next-token prediction objectives \citep{lai2025hierarchical}, more recent findings reveal that intermediate layers can outperform final ones in extractability, depending on the task and readout mechanism \citep{skean2024representation}. Additionally, early layers prioritize surface-level or lexical features, intermediate layers support syntactic structure, and later layers encode increasingly semantic or task-specific signals, though this profile varies by architecture and analytic method \citep{jawahar2019bert, rogers2020bertology}.

One of the primary methods for localizing such information is probing, wherein lightweight classifiers are trained to decode linguistic properties from embeddings \citep{conneau2018cram, tenney2019context}. Probing across layers provides a practical map of when linguistic traits become linearly decodable. However, outcomes depend heavily on probe design, capacity, regularization, and controls, making it difficult to infer representational structure directly from classification scores \citep{pimentel2020information}. Structural probing partially addresses this by aligning embedding-space distances with syntactic structure, connecting decodability to geometric relationships \citep{hewitt2019structural}.

To address these limitations, geometry-based analyses characterize representational layout independent of classifier performance. Techniques such as representational similarity analysis (RSA) and centroid-based separation measure how embeddings cluster or diverge across layers \citep{kriegeskorte2008representational, choenni2019blackbox}, offering complementary insight into the model's internal organization. Such measures are particularly important given known shifts in geometry across depth, including anisotropy, context specificity, and feature superposition that affect interpretability and separability \citep{ethayarajh2019contextual, elhage2022toy}. Moreover, dimensionality and sparsity vary across architectures, meaning that top dimensions or principal components may encode overlapping features in different ways \citep{gurnee2023neurons}. These factors make it essential to jointly consider linear decodability and geometric separation when analyzing category representations.

Building on prior work showing that standard language (e.g., conventional) exhibits stronger model-human alignment than figurative categories (e.g., idioms, Gen Z slang) \citep{bollepally2026figurative}, we ask whether these behavioral differences reflect internal category encoding. While probing and geometry may diverge due to subspace encoding or anisotropic structure \citep{ethayarajh2019contextual, elhage2022toy, pimentel2020information}, their joint analysis offers a mechanistic lens into representational divergence. Treating human judgments as a semantic reference, we focus on internal structure rather than output behavior to ask: (1) Do models differentiate linguistic categories internally? (2) Where in the hierarchy is this most pronounced? and (3) how do these trajectories differ across architectures?

\subsection{Overview}
Focusing on the distinction between conventional and idiomatic sentences, a core for human-aligned category-based semantic processing \citep{bridges2015:129, Yamauchi2009}, we analyze two 1B-parameter pre-trained models (Llama-3.2-1B and Gemma-3-1B-PT) using layer-wise probing classification and geometric distance analyses over hidden representations, and additionally, we used Principal Component Analysis (PCA) to identify the most informative embedding dimensions at each layer.

\begin{figure*}[h!]
\centering
% --- Row 1: Study 1 ---
\begin{subfigure}[t]{0.48\textwidth}
  \centering
  \includegraphics[width=\linewidth]{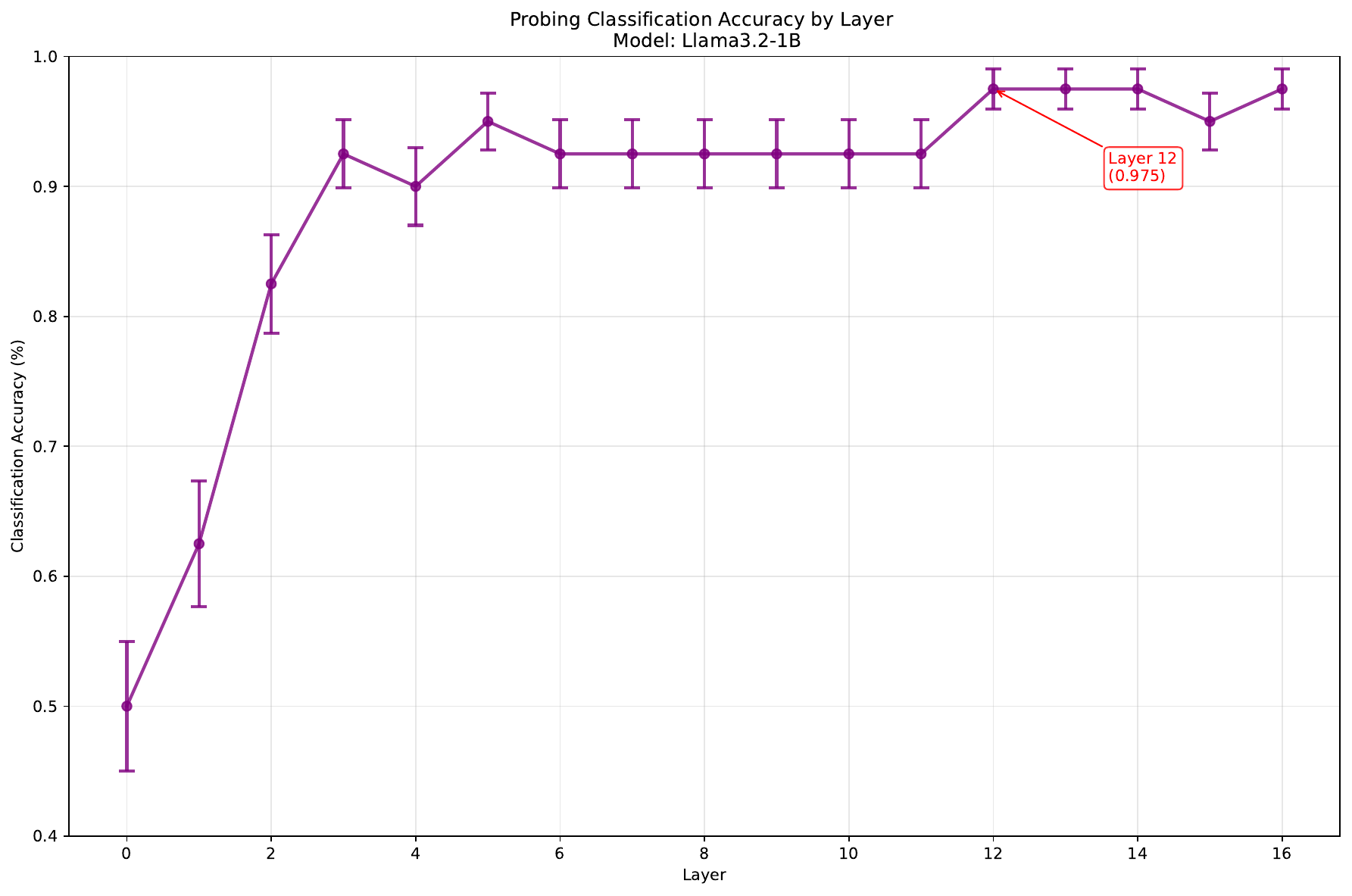}
  \caption{}
  \label{fig:probing-L}
\end{subfigure}
\hfill
\begin{subfigure}[t]{0.48\textwidth}
  \centering
  \includegraphics[width=\linewidth]{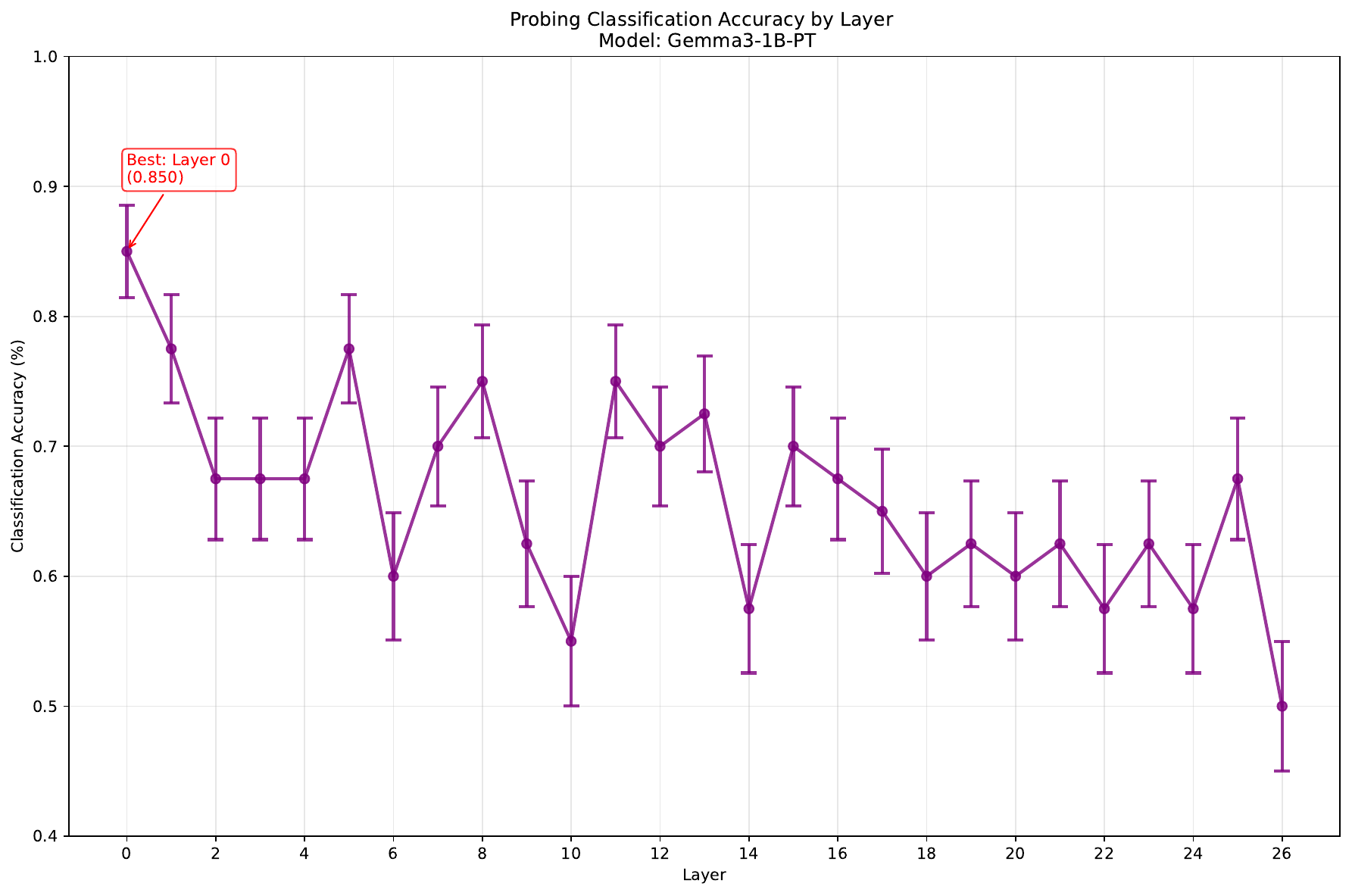}
  \caption{}
  \label{fig:probing-G}
\end{subfigure}
\caption{The classification accuracy per layer of logistic regression classifier using top 15 original embeddings per layer for models (a) Llama-3.2-1B and (b) Gemma-3-1B-PT.}
\label{fig:prob}
\end{figure*}

The results reveal qualitatively distinct representational profiles across models. Llama-3.2-1B exhibits a canonical deepening pattern in which category information is weakly expressed in early layers, becomes robustly linearly decodable in mid-to-late layers, and is stably maintained through depth via a relatively consistent set of embedding dimensions (Figure \ref{fig:probing-L}). In contrast, Gemma-3-1B-PT shows strongest category accessibility at the input and early layers, followed by progressive attenuation of both linear decodability and geometric separation in deeper layers, indicating a redistribution of category-relevant structure across the representation space (Figure \ref{fig:probing-G}). Together, these findings provide a mechanistic account of how different model architectures organize linguistic category information across layers and offer a potential explanation for cross-model differences in alignment with human judgments on figurative language tasks.

\section{Method}

To evaluate how sentence categories are differentiated within internal representations, layer-wise embeddings from large language models were analyzed across the network. 

\subsection{Models}
Two pre-trained large language models, Llama-3.2-1B and Gemma-3-1B-PT, were selected to enable a comparative analysis of layer-wise representational dynamics across different model architectures. Both models were acquired from the Hugging Face platform.

\subsection{Dataset}
The dataset consisted of 40 sentences, divided into two categories: conventional and idiomatic. Each category contained 20 sentences. Conventional sentences comprised straightforward, commonly used expressions encountered in everyday language (e.g., \textit{"the dog is outside"}), whereas idiomatic sentences involved figurative constructions in which meaning emerges from the sentence as a whole rather than from the literal interpretation of individual words (e.g., \textit{"bite the bullet"}). This dataset is extracted from \citep{bollepally2026figurative}.

\subsection{Procedure}

Each sentence was independently processed by both models, with hidden-state representations extracted from every layer. For each sentence and layer, token-level embeddings were mean-pooled to obtain a single fixed-length vector. Llama-3.2-1B yields 2048-dimensional representations per layer, whereas Gemma-3-1B-PT yields 1152-dimensional representations. To identify the most impactful dimensions at each layer, Principal Component Analysis (PCA) was applied independently using sentence-level embeddings from all sentences. PCA was used strictly to select dimensions explaining the maximum variance, rather than as a representational transformation.

\textbf{Principal Component Analysis:} For each model, the number of principal components required to explain at least 70\% of the variance was first determined separately for each layer. The maximum number of components required by any layer to reach this threshold was then selected and applied uniformly across all layers, resulting in 20 principal components per layer for Llama-3.2-1B and 18 principal components per layer for Gemma-3-1B-PT. The PCA loadings associated with the selected components were then used to rank the original hidden dimensions by their contribution. The top 15 dimensions per layer with the highest absolute loadings were retained. These dimensions were layer-specific, and the corresponding original embedding values were extracted for all sentences for subsequent analyses.

\textbf{Analysis 1: Probing via Classification}

A probing analysis was conducted to assess category separability at each layer using the selected embedding dimensions. For each layer, the top 15 embedding values per sentence and its category information were served as input features, with sentence category (conventional vs. idiomatic) as the target label. A logistic regression classifier was trained and evaluated using 5-fold cross-validation.

To assess whether classification performance exceeded chance levels, a permutation test was conducted by randomly shuffling sentence category labels and recomputing classification accuracy. Statistical significance was evaluated using p-values derived from the permutation distribution.

\textbf{Analysis 2: Geometric Separation via Centroid Distances}

As a complementary geometric analysis, centroid representations were computed separately for each sentence category per hidden dimension at each layer. For a given layer, the centroid was obtained by averaging each selected embedding dimension across all sentences within a category. Euclidean distances were then computed between category centroids to quantify representational dispersion and separation. These centroid-based Euclidean distances provide a layer-wise measure of geometric separability between conventional and idiomatic sentence representations.

\section{Results}

The analysis examines the layer-wise patterns between consecutive layers in two models (Llama-3.2-1B and Gemma-3-1B-PT) when processing conventional, idiomatic sentences, and analyzes their differences. The findings reveal distinct processing patterns and representational differences between the two sentence types across different stages of layers.

\textbf{A. High-Variance Embedding Dimensions Across Layers}
\begin{figure*}[h!]
\centering
% --- Row 1: Study 1 ---
\begin{subfigure}[t]{0.48\textwidth}
  \centering
  \includegraphics[width=\linewidth]{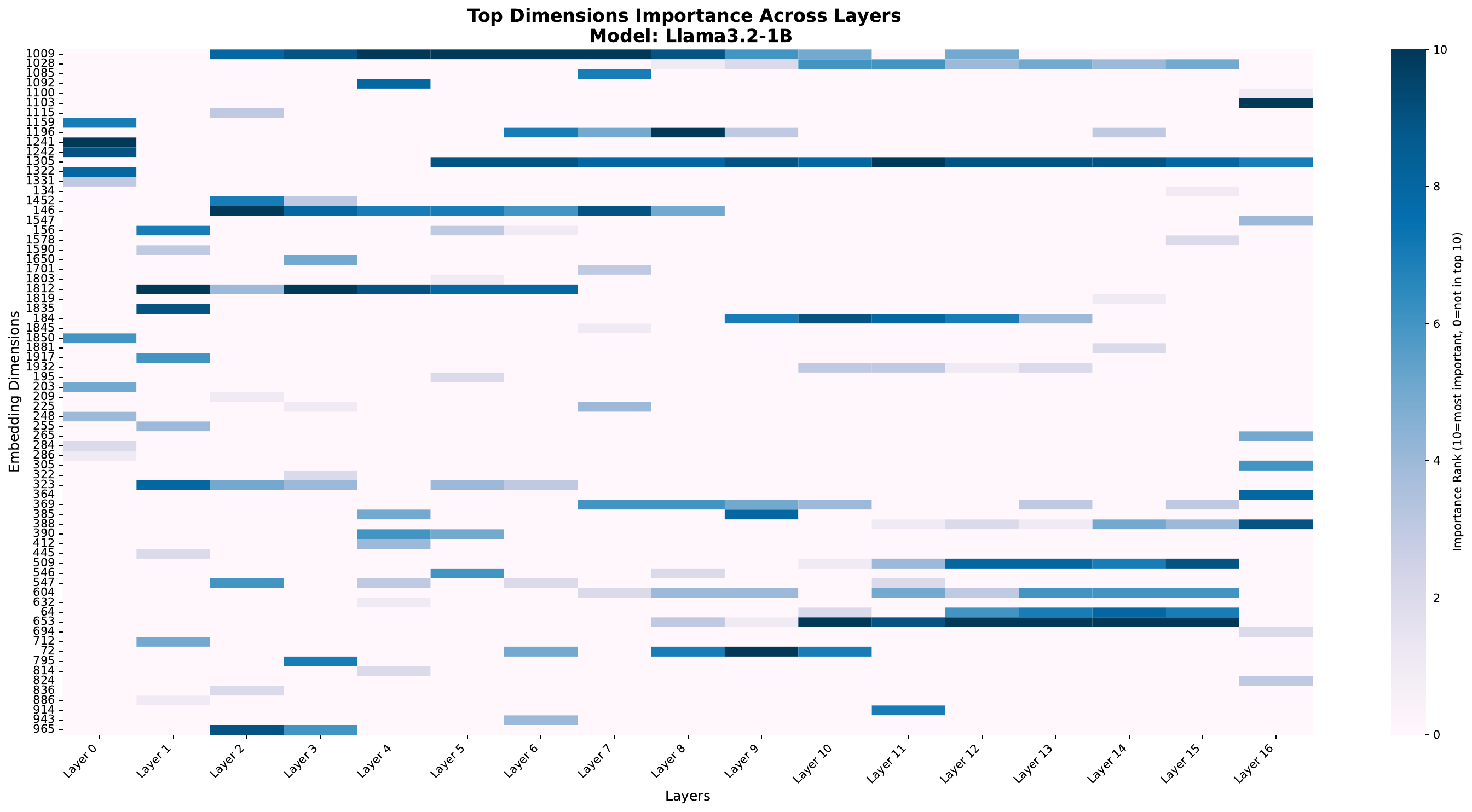}
  \caption{}
  \label{fig:dim-L}
\end{subfigure}
\hfill
\begin{subfigure}[t]{0.48\textwidth}
  \centering
  \includegraphics[width=\linewidth]{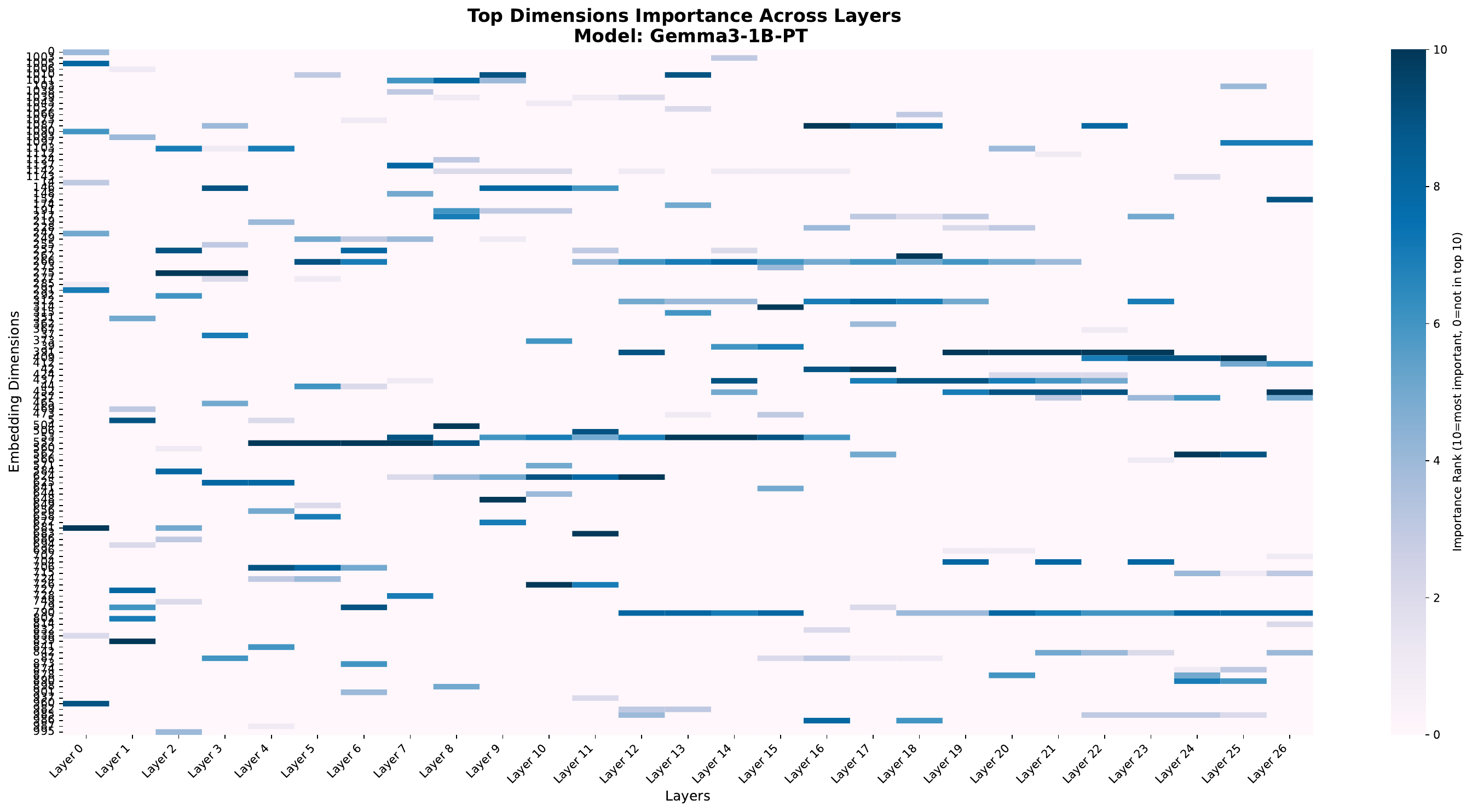}
  \caption{}
  \label{fig:dim-G}
\end{subfigure}
\caption{The prominence of top 15 embedding dimensions per layer for models (a) Llama-3.2-1B and (b) Gemma-3-1B-PT}
\label{fig:dim}
\end{figure*}

% Figure \ref{fig:dim} presents the layer-wise distribution of top-ranked original embedding dimensions derived from principal components explaining over 70\% of total variance per layer in each model, with dimension importance defined by global variance contribution rather than category-specific discrimination. 

In Llama-3.2-1B, a relatively sparse and structured pattern is observed: a small subset of embedding dimensions repeatedly appears among the top-ranked dimensions across multiple layers, particularly in the early and late stages of the network.  In contrast, Gemma-3-1B-PT exhibits a more heterogeneous pattern of dimension importance across layers. The set of top-ranked dimensions varies substantially from layer to layer, with fewer dimensions recurring consistently across depth. This dispersion reflects a more distributed organization of representational variance.

\textbf{B. Probing Performance Across Layers}

In Llama-3.2-1B, classification accuracy increases sharply from early to mid layers and remains consistently high thereafter, reaching its maximum between layers 12 and 16 (accuracy = 0.975). This sustained plateau indicates that sentence category information becomes robustly and stably linearly decodable in later layers.

In contrast, Gemma-3-1B-PT attains its highest classification accuracy at the input layer (layer 0; accuracy = 0.85), followed by a general decline across subsequent layers. Although moderate accuracy is intermittently recovered in some mid layers, overall performance degrades toward deeper layers, approaching chance levels at the final layer. 

Permutation tests confirmed that probing accuracies at the peak-performing layers for both models significantly exceeded chance levels ($\textit{p} < 0.05$). 

\textbf{C. Layer-wise Geometrical Category Separation}
\begin{figure*}[h!]
\centering
% --- Row 1: Study 1 ---
\begin{subfigure}[t]{0.48\textwidth}
  \centering
  \includegraphics[width=\linewidth]{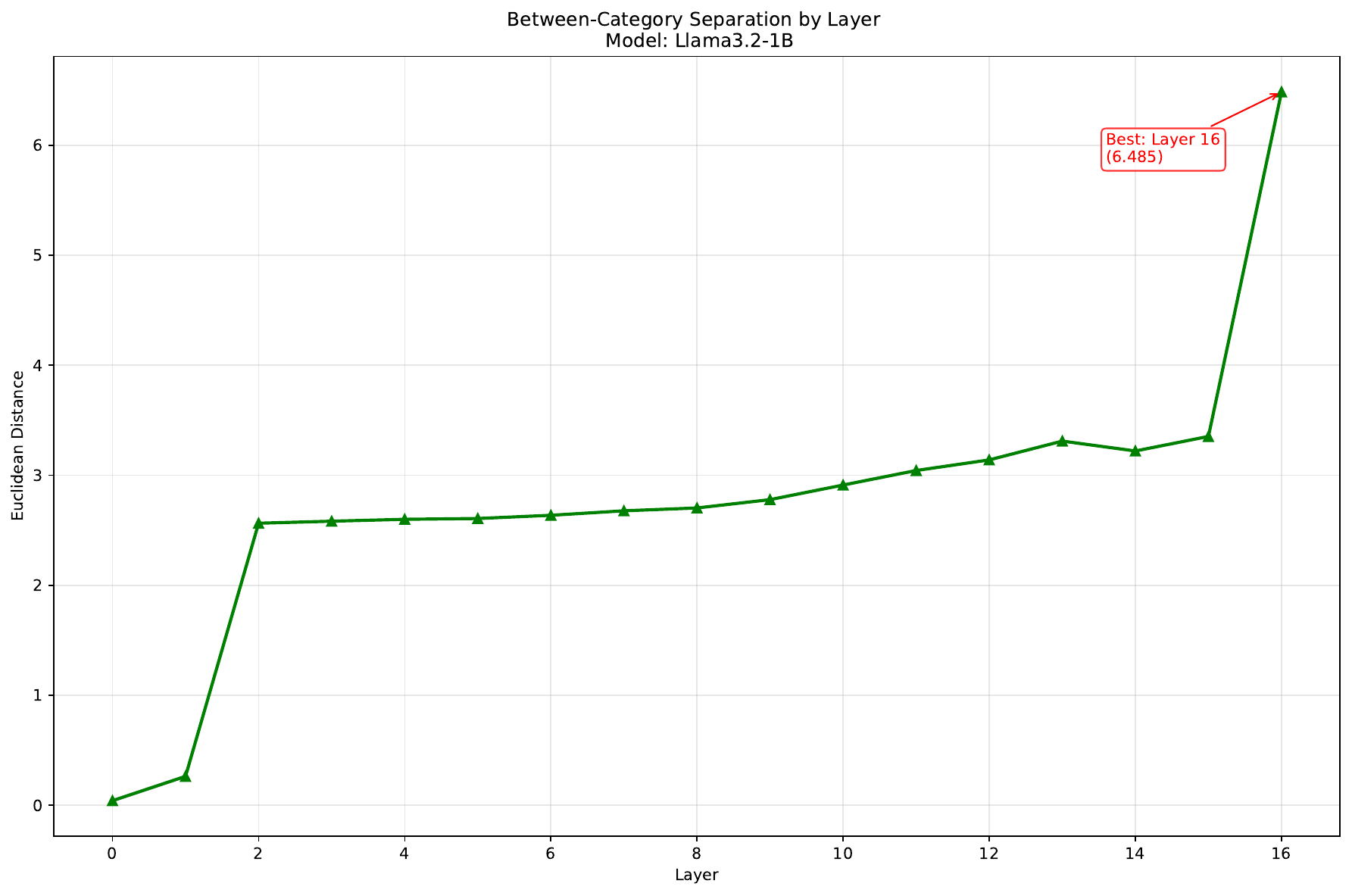}
  \caption{}
  \label{fig:cd-L}
\end{subfigure}
\hfill
\begin{subfigure}[t]{0.48\textwidth}
  \centering
  \includegraphics[width=\linewidth]{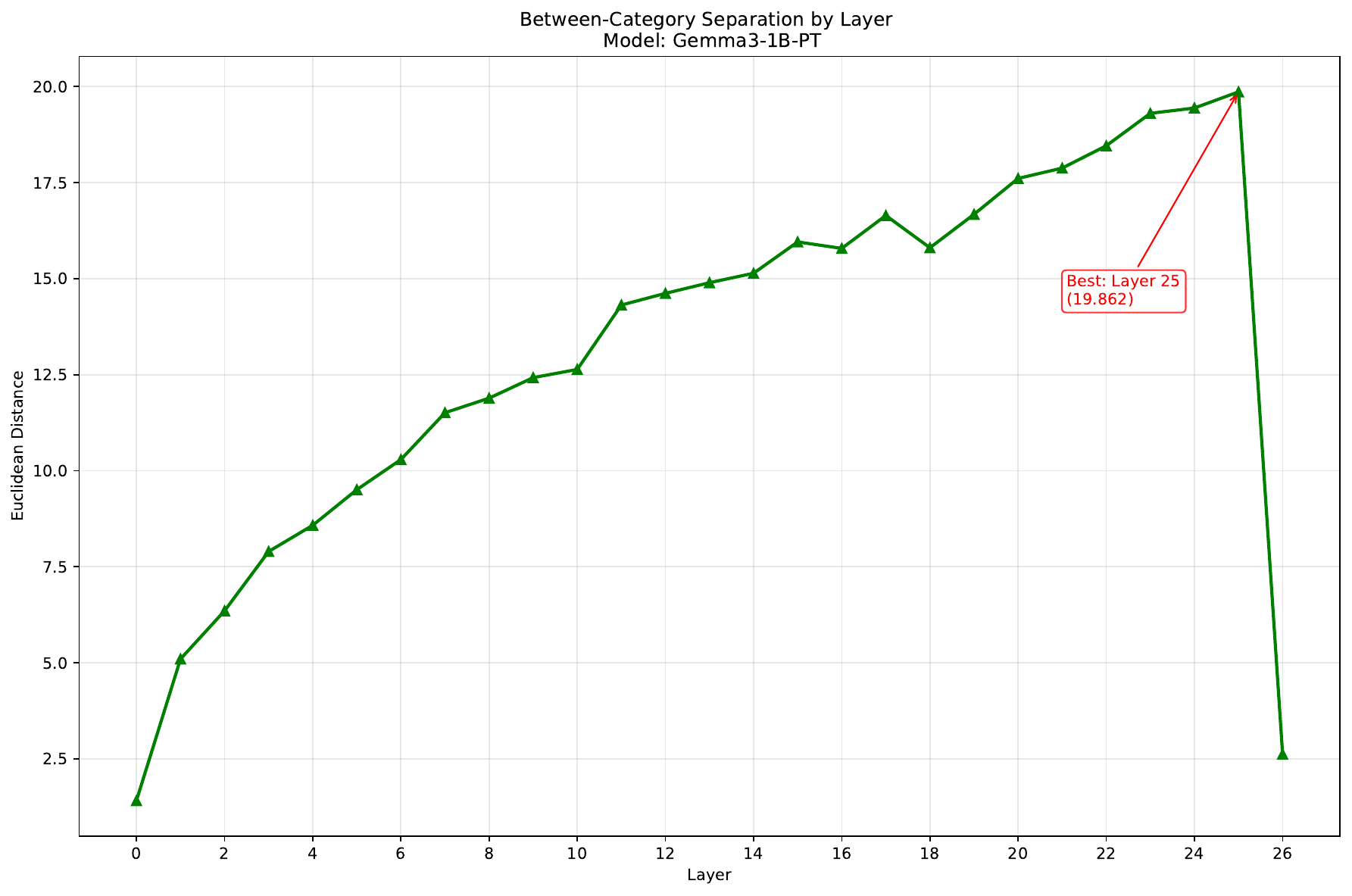}
  \caption{}
  \label{fig:cd-G}
\end{subfigure}
\caption{The Euclidean distance between conventional and idiomatic sentences per layer for models (a) Llama-3.2-1B and (b) Gemma-3-1B-PT.}
\label{fig:cd}
\end{figure*}

In Llama-3.2-1B, the Euclidean distance between categories increases sharply in the early layers, followed by a relatively stable plateau with only marginal incremental changes across intermediate layers, and reaches its maximum at the final output layer (Layer 16). This pattern indicates that geometric differentiation between sentence categories emerges early and is subsequently maintained and modestly amplified through depth.

In contrast, Gemma-3-1B-PT exhibits a more gradual and sustained increase in between-category geometrical distance from early to mid layers, accompanied by fluctuations in intermediate layers that nevertheless remain above early-layer values. The distance reaches its maximum at the final hidden layer (Layer 25), but then drops sharply at the final output layer, approaching levels observed in the initial hidden layers. This late-stage contraction suggests a reduction in geometric separation between sentence categories immediately before output generation.

\section{Discussion}

For Llama-3.2-1B, the probing accuracy rose sharply in early to mid layers and plateaued at final layers, indicating that linguistic trait information becomes linearly accessible mid-depth and remains stable thereafter. This stability is reinforced by the repeated use of top embedding dimensions across layers, suggesting that task-relevant features are preserved and reweighed rather than discarded \citep{skean2024representation, conneau2018cram}. Geometric analysis showed a parallel trend: Euclidean distance between categories peaked at the final output layer, aligning with probing results and confirming that category information is both decodable and structurally embedded in the representational space (\citep{kriegeskorte2008representational, choenni2019blackbox}. Together, these findings reflect a canonical deepening trajectory in which separable structure emerges in mid-layers and is refined toward the final layer.

In Gemma-3-1B-PT, probing accuracy peaked at the input layer and declined steadily across depth, reaching chance levels by the final layer, indicating that category information is most accessible early but becomes increasingly non-linearly encoded. This is consistent with architectures that redistribute information across dimensions without preserving linearly decodable structure \citep{elhage2022toy, gurnee2023neurons}. The absence of recurring top embedding dimensions further supports this continuous reparameterization. Geometrically, centroid distance peaked at Layer 25 but collapsed at the final layer, suggesting that mid-depth separation is not stably maintained. The mismatch between geometric separation and low probing accuracy implies that category information may remain embedded but not linearly accessible, likely due to high variance or misaligned subspaces \citep{pimentel2020information, choenni2019blackbox}.

Although both the models have identical parameter counts, their internal representations differ substantially. These differences arise not only from architectural choices such as layer normalization, attention routing, and embedding dimensionality, but also from training objectives, initialization schemes, and representational priors. As a result, even with equal capacity, models may organize linguistic information in fundamentally different ways, weakening the alignment of linguistic interpretation between human and model representational spaces  \citep{bollepally2026figurative}.  These findings reinforce the need to analyze internal representational geometry, examining not just behavioral outputs but semantic space alignment when evaluating how closely human and model language interpretation correspond \citep{bollepally2026figurative,  skean2024representation}.

\section{Limitations and Future Work}
While this study offers insight into layer-wise representational organization across architectures, it is constrained by a small dataset of 40 sentences and focuses solely on linear probes and centroid-based geometric analysis. Future work could broaden this framework by incorporating nonlinear probing methods and information-theoretic metrics to deepen linguistic characterization with a broader dataset. Also, the direct link of internal representational metrics to humans' would help understand the extent of cognitive alignment with humans.

\section{Conclusion}

Using a layer-wise approach, this study shows that while both Llama-3.2-1B and Gemma-3-1B-PT encode linguistic category distinctions internally, they do so through distinct representational processes. The findings suggest that improving layer-wise consistency, particularly by preserving stable, decodable subspaces may enhance a model’s ability to retain semantic distinctions across depth, highlighting a promising direction for architectural and training improvements beyond scaling.

\bibliographystyle{unsrt}
\bibliography{reference}
\clearpage
\onecolumn
\section*{Supplementary Materials}

To determine which embedding dimensions have a greater influence on category differentiation, Cohen’s d was calculated across all layers by comparing the embedding values of each category. The analysis shows Llama-3.2-1B superior discrimination ability, with the average Cohen's d value reaching 1.94 compared to Gemma-3-1B-PT's maximum of 1.24, indicating that Llama has learned much more robust representations for distinguishing conventional vs idiomatic expressions.

\begin{table*}[h]
\centering
\small
\setlength{\tabcolsep}{10pt}
\begin{tabular}{l l l}
\toprule
\textbf{Dimension no.} & \textbf{Avg. Cohen's d}  \\
\midrule
1305 & 1.9380 \\ 
1103 & 1.7928 \\
509 & 1.7324 \\
653 & 1.6987 \\
1812 & 1.6622 \\
\bottomrule
\end{tabular}
\caption{The most impactful dimensions for differentiating sentence categories across all layers in Llama-3.2-1B}
\label{tab:sentences}
\begin{tabular}{l l l}
\toprule
\textbf{Dimension no.} & \textbf{Avg. Cohen's d}  \\
\midrule
504 & 1.2418 \\ 
552 & 1.1128 \\
275 & 1.0642 \\
683 & 0.9727 \\
648 & 0.8985 \\
\bottomrule
\end{tabular}
\caption{The most impactful dimensions for differentiating sentence categories across all layers in Gemma-3-1B-PT}
\label{tab:sentences}
\end{table*}

\clearpage
% \onecolumn
% \appendix
% \section{Complete Sentence Set}
% \input{Appendix_All_240_Sentences}
% \clearpage
% \section{Interpretive Questions}
% \input{Questions}

\end{document}